\documentclass[pdflatex,sn-mathphys-num]{sn-jnl}

\usepackage{graphicx}%
\usepackage{multirow}%
\usepackage{amsmath,amssymb,amsfonts}%
\usepackage{amsthm}%
\usepackage{mathrsfs}%
\usepackage[title]{appendix}%
\usepackage{xcolor}%
\usepackage{textcomp}%
\usepackage{manyfoot}%
\usepackage{booktabs}%
\usepackage{algorithm}%
\usepackage{algorithmicx}%
\usepackage{algpseudocode}%
\usepackage{listings}%

\usepackage{soul}

\usepackage{draftwatermark}
\SetWatermarkText{Under-review}
\SetWatermarkScale{0.5} 

\theoremstyle{thmstyleone}%
\theoremstyle{thmstyletwo}%

\theoremstyle{thmstylethree}%

\begin{document}

\title[Article Title]{Bias in Filter Feature Selection Evaluation: A Meta-Analysis of Datasets, Baselines, and Experimental Design Choices}

\author*[1,3]{\fnm{Malick} \sur{Ebiele}}\email{malick.ebiele@adaptcentre.ie}

\author[2,3]{\fnm{Malika} \sur{Bendechache}}\email{malika.bendechache@universityofgalway.ie }

\author[1,3]{\fnm{Rob} \sur{Brennan}}\email{rob.brennan@ucd.ie}

\affil*[1]{\orgdiv{Computer Science}, \orgname{University College Dublin}, \orgaddress{ \city{Dublin}, \state{Dublin}, \country{Ireland}}}

\affil[2]{\orgdiv{Computer Science}, \orgname{University of Galway}, \orgaddress{ \city{Galway}, \state{Galway}, \country{Ireland}}}

\affil[3]{\orgname{ADAPT Centre}, \orgaddress{\city{Dublin}, \state{Dublin}, \country{Ireland}}}


\abstract{

\textbf{Background:} Since 1990 many feature selection methods have been proposed across heterogeneous applications. To validate the usefulness of a new method, it needs to be compared against at least one baseline method from the existing literature on a feature selection task using at least one dataset. Recent developments in tabular Deep Learning (DL) and data valuation in Machine Learning (ML) suggest that the evaluation of new methods, algorithms, and models may be consciously or unconsciously biased. We hypothesise that a similar trend exists in feature selection (FS), particularly in filter feature selection (FFS). The aim of this study is therefore to examine FFS studies to identify factors that influence the evaluation and that might consist entry point for biases in order to recommend stronger principles for FFS evaluation. 

\textbf{Methods:} We analyse a sample of 28 high profile FFS studies published between 1994 and 2025. The analysis provides reflections on how to examine FFS studies, highlights lessons learned throughout the process, and gives five evidence-based recommendations for future FFS evaluation. 

\textbf{Results:} Multivariate Linear Regression analysis achieved a score of $R^2=0.33$. It means that 33\% of the variance in the performance of new methods against chosen baselines (win rate) is explained by the number of datasets (\#Datasets), the number of baselines (\#Baselines), and the number of new methods (\#NewMethods).

\textbf{Discussion:} $R^2=0.33$ is considered medium explanation; which is promising given that this is the first such study. The medium explanation result is due to the fact that win rate is influenced by additional factors such as the maturity of the feature selection domain, the type of datasets and baselines, and the simplicity of the regression model used to explain the relationship. 




}




\maketitle


\section{Introduction}\label{chap6:sec:Introduction}

Over the last four decades, the number of publications related to feature selection has increased rapidly, going from about 3000 in 1990 to over 30000 in 2020 in the IEEE and ACM databases alone \cite{theng_feature_2024}. Most of those studies are application and validation studies; meaning they do not introduce a new method. 

\textbf{Feature selection (FS) methods in machine learning (ML) consist of ways to select a subset of $k$ features from the original feature set $\mathcal{F}$ such that $k<|\mathcal{F}|$, for example to reduce computational complexity, while maintaining a competitive predictive performance} \cite{guyon_introduction_2003, jiao_survey_2023}. There are three primary types of FS methods: \textbf{filters, wrappers,} and \textbf{embedded} \cite{guyon_introduction_2003, jiao_survey_2023, moslemi_tutorial-based_2023, theng_feature_2024}. During the period 1990-2020, 129 new methods (or extensions of an existing method) have been introduced including 55 filter, 49 wrapper, and 25 embedded methods \cite{theng_feature_2024}\footnote{For more detail on each primary type of FS method, please refer to any of the following studies: \citet{guyon_introduction_2003, jiao_survey_2023, moslemi_tutorial-based_2023}, and \citet{theng_feature_2024}. For instance, \citet{jiao_survey_2023} provides a comparative analysis of the three primary FS types in terms of relative predictive performance, computational cost, and generalisation [of the selected features] to other models.}. 

When to comes to the comparative analysis of FS method effectiveness in terms of reported predictive performance, the results are not always conclusive. It is well-known in the feature selection community that wrappers are more ``likely'' to outperform embedded methods, which in turn tend to outperform filters. We used the term ``likely'' because filters sometimes surpass both wrappers and embedded methods \cite{bommert_benchmark_2022, mostert_filter_2018, li_new_2020}. Predicting in advance which method will be superior remains very difficult. This uncertainty existed 23 years ago, which led \citet{guyon_introduction_2003} to recommend investigating all types of feature selection methods. They emphasised that the only constraints should be computational resources and time. 
Additionally, in the processing of NeurIPS 2003 competition results, \citet{guyon_result_2004} found that several top participants used simple methods such as KBest (a filter) with correlation coefficients as relevance estimators. \citet{guyon_result_2004} also noted that top participants used combinations of relevance estimators, feature search algorithms (e.g., KBest, mRMR), and classifiers. In short, achieving high predictive performance requires the right combination of a relevance estimator, a search algorithm, and a classifier. It is very difficult to know this combination in advance, confirming the exploratory nature of feature selection. This complicates FFS evaluation for new methods as the search space is potentially very large and the stopping conditions or best practices for robust evaluation are not well documented or often studied in their own right. This paper aims to address this gap.

This study is also inspired by recent developments in ML evaluation, especially in tabular Deep Learning (DL) and data valuation in ML. For instance, in 2022, \citet{grinsztajn_why_2022} and \citet{shwartz-ziv_tabular_2022} showed that tree-based models outperform specialised DL models on tabular datasets. \citet{shwartz-ziv_tabular_2022} went further, showing that tree-based models outperform tabular DL models on datasets used in the original studies that introduced those tabular DL models. This raises questions about whether evaluation choices influenced the relative performance of the compared methods. This, for example, can be seen in the limited hyperparameter finetuning of tree-based models \cite{cherepanova_performance-driven_2024}, even though tree-based models are significantly faster to train than tabular DL models and hence easier to finetune \cite{grinsztajn_why_2022, borisov_deep_2024}.

Evidence of poor evaluation practice is demonstrated in a recent 2024 tabular DL study. \citet{cherepanova_performance-driven_2024} introduced Deep Lasso, claiming it outperforms Random Forest (RF) and eXtreme Gradient Boosting (XGB) models—contradicting earlier findings by \citet{grinsztajn_why_2022} and \citet{shwartz-ziv_tabular_2022}. However, their methodology raises several concerns. 
First, the hyperparameter spaces used for finetuning RF and XGB are smaller than those considered by \citet{grinsztajn_why_2022}, introducing possible bias. Second, RF and XGB were used as feature extractors for Multi-Layer Perceptron (MLP) and FT-Transformer\footnote{FT-Transformer stands for Feature Tokeniser + Transformer. It was originally proposed by \citet{gorishniy_revisiting_2021} as a tabular deep learning model.}, rather than as independent baselines; in contrast, prior studies have either used tree-based models as standalone predictors or input features from other models. Third, direct comparison is hindered because \citet{cherepanova_performance-driven_2024} used different datasets and performance evaluation metrics. Fourth, they considered only 12 datasets versus 45 in \citet{grinsztajn_why_2022}, used distinct aggregation methods, and omitted leading tree-based models previously shown to outperform tabular DL, such as CatBoost, LightGBT, and HistGradientBoosting \cite{grinsztajn_why_2022, borisov_deep_2024, shwartz-ziv_tabular_2022, gorishniy_revisiting_2021}.

In 2023, \citet{jiang_opendataval_2023} performed a benchmark analysis of data valuation techniques in ML and concluded that no single method is uniformly superior across all evaluation metrics and tasks (i.e., Noisy label detection, Noisy feature detection, Point removal, and Point addition). When examining those individual data valuation studies, the proposed method is shown to be clearly superior to the baselines. This is in line with observations in the tabular DL domain by \citet{grinsztajn_why_2022}, \citet{shwartz-ziv_tabular_2022}, and \citet{borisov_deep_2024}. This suggests a lack of robust evaluation of newly proposed methods, algorithms, or models, which can lead to unrealistic conclusions. The new methods, algorithms, or models are often presented in ways that make them appear clear winners, possibly reflecting broader publication norms that value clear performance improvement.

Therefore, we hypothesise that a similar trend exists in FS, particularly in FFS, and aim to study factors that influence the win rate of new methods in FFS. The research question is the following: \textbf{To what extent do the number of datasets, the number of baselines, and the number of new methods influence the win rate of newly proposed filter feature selection methods, as reported in the literature?} To address this research question, a sample of 28 studies, published between 1994 and 2025, have been selected and analysed. 

The remainder of this paper is structured as follows. 
Section \ref{chap6:sec:Methodology} describes the methodology. Section \ref{chap6:sec:Analysis of the Current Feature Selection Evaluation Landscape} analyses the feature selection evaluation landscape and provides targeted recommendations. Section \ref{chap6:sec:Lessons Learned} highlights lessons learned during our analysis. General recommendations are in section \ref{chap6:sec:General Recommendations}. Section \ref{chap6:sec:Summary and Conclusion} provides the summary and conclusion.

\section{Methodology}\label{chap6:sec:Methodology}

This section describes the methodology used to address the research question stated in section \ref{chap6:sec:Introduction} above.

\subsection{Contribution Evaluation Metrics}\label{chap6:subsec:Metrics}
To achieve our objective of using a common framework to assess the respective contributions of the experimental results reported in the literature, we defined two new metrics: win rate and success rate.

\textbf{The win rate is defined as the number of datasets the proposed methods (considered as a single ensemble of methods) outperform all the baselines divided by the total number of reported datasets; the result is multiplied by 100.} By reported datasets, we mean the datasets included in the comparison of the proposed methods with the baselines using a performance metric such as accuracy. The reason is that some studies only report a subset of the datasets mentioned or described \cite{battiti_using_1994, kwak_input_2002, zhang_compact_2014, huang_multilabel_2020, yuan_cscim_fs_2023, yuan_feature_2025} or use different metrics on different subsets of the included datasets \cite{yuan_feature_2024, zhang_compact_2014}. Other studies report detailed results, e.g. different results per classifier such as Linear Support Vector machine Classifier (LinearSVC) or RF \cite{ihianle_minimising_2024, zhao_maximum_2019, jo_improved_2019, peng_feature_2005}. There are studies which report aggregated results, but using heterogeneous aggregation techniques \cite{ihianle_minimising_2024, bugata_aspects_2019, bommert_benchmark_2022, yuan_feature_2025, zhang_feature_2020, bennasar_feature_2015, zhao_maximum_2019}. However, \textit{details results using different values of $k$ (the number of selected features) are considered the same configuration} because this exercise consists of finetuning of the number of selected features to determine the best value according to the experimental setup. In the above scenarios, we reported in table \ref{tab:Sample FS studies} only the configurations where the proposed methods achieved the highest win rate, and we used the same configuration for the success rate.

\textbf{The success rate is defined as the number of datasets the proposed methods (considered as a single ensemble of methods) outperform the original data using the entire feature set (all-feature) divided by the total number of reported datasets; the result is multiplied by 100.} The success rate calculation follows the same procedure as the win rate (as described above), but instead of the baselines, the proposed methods are compared to the original data using the entire feature set (all-feature).

The win rate and success rate are not designed to measure statistical superiority in the classical inferential sense. Rather, they serve as descriptive meta-evaluation indicators that quantify how frequently newly proposed methods are reported to outperform competing baselines or the all-feature setting within their respective studies. Their purpose is to characterise reporting patterns across the literature under heterogeneous experimental conditions.

\subsection{Study Selection}\label{chap6:subsec:Study Selection}
The objective is to identify the set of the most common families of FFS methods and baselines evaluated in the literature on high-dimensional datasets (in order to observe the evaluation methods used). 
The initial study selection of FFS methods started with three author-selected studies from previous work. i) A data-centric systematic review of feature selection methods by \citet{li_feature_2017}, ii) the analysis of NeurIPS 2003 feature selection challenge results presented by \citet{guyon_result_2004}, and iii) a filter feature selection benchmark paper by \citet{bommert_benchmark_2022}. We only focus on methods that are regularly used or mentioned as baselines for high-dimensional feature selection tasks or extensions of those methods. We then traced the baselines included in those studies to the earliest mention (we could find) of each method. The final study selection was performed manually guided by our experience in feature selection and two additional review papers: i) one recent systematic feature selection review by \citet{theng_feature_2024} and ii)  \citet{guyon_introduction_2003}'s famous introduction to variable and feature selection. 


As a result, 28 sample studies have been included in our analysis. We acknowledge that the 28 sample studies represent a small fraction of all published FS studies. However, the relevant population for this study is not all FS publications, but specifically FFS studies that introduce a new method and provide a comparative evaluation against baselines. Based on \citet{theng_feature_2024}, this population is estimated at approximately 55 new FFS methods over three decades, making our sample of 28 a substantial proportion (51\%) of the target population. Additionally, we believe that the 28 sample studies are a good indicator of quality, as most have been published at top-tier conferences and journals.



\subsection{Data Extraction}\label{chap6:subsec:Data Extraction}
To compute the win and success rates, we first need to extract the number of datasets and baselines, and identify all the new methods proposed in each sample study. The extraction was performed manually by reading each sample study. We also extracted from each study the reference, type (conference or journal), year [of publication], title, and venue. All the extracted data is presented in table \ref{tab:Sample FS studies} below.

\subsection{Data Analysis}\label{chap6:subsec:Data Analysis}

After data extraction, Linear Regression (LR) was used as an exploratory tool to analyse the association between win rate and three variables: number of datasets (\#Datasets), number of baselines (\#Baselines), and number of new methods (\#NewMethods). Seven regression models have been trained. Three univariate regression models, one for each of the three variables: \#Datasets, \#Baselines, and \#NewMethods (see figure \ref{chap6:fig:univariate correlation between Win Rate and Sample data}). Three bivariate regression models have been trained, one for \#Datasets and \#Baselines, \#Datasets and \#NewMethods, and \#Baselines and \#NewMethods, respectively (see figure \ref{chap6:fig:bivariate correlation between Win Rate and Sample data}). Finally, one for all three variables together (see figure \ref{chap6:fig:correlation between Win Rate and Sample data}). On figures \ref{chap6:fig:bivariate correlation between Win Rate and Sample data} and \ref{chap6:fig:correlation between Win Rate and Sample data}, the x-axis represents one-component Principal Component Analysis (PCA) with the corresponding explained variance. This allows us to simplify the visualisations instead of three- and four-dimension graphs. 

\textbf{The regression models have been evaluated using the coefficient of determination (R-squared or $R^2$) and the Root Mean Square Error (RMSE) metrics. $R^2\in[0;1]$ with 1 being the optimal value. RMSE $\in[0;\infty)$ with 0 the optimal value.} $R^2$ and RMSE formulae are described in equations (\ref{chap6:eq:R-squared}) and (\ref{chap6:eq:RMSE}), respectively.

\begin{equation}
R^2 (y, \hat{y}) = 1 - \frac{\displaystyle\sum_{i=1}^{n} (y_i - \hat{y}_i)^2}{\displaystyle\sum_{i=1}^{n} (y_i - \bar{y})^2} = 1 - \frac{SS_{\text{res}}}{SS_{\text{tot}}}
\label{chap6:eq:R-squared}
\end{equation}
Where $n$ is the number of observations, $y_i$ the actual value, and $\hat{y}i$ is the predicted value. $\bar{y} = \displaystyle\frac{1}{n} \sum{i=1}^{n} y_i$ is the sample mean. $SS_{\text{res}}$ is the sum of squares of residuals. $SS_{\text{tot}}$ is the total sum of squares (proportional to the variance of the data, i.e. $SS_{\text{tot}} = (n-1)\times\operatorname{Var}(y)$  ).

\begin{equation}
\operatorname{RMSE} (y, \hat{y}) = \sqrt{\frac{1}{n} \displaystyle\sum_{i=1}^{n} (y_i - \hat{y}_i)^2}
\label{chap6:eq:RMSE}
\end{equation}

\begin{table}[!htp]\centering
\tiny
\caption{Sample FFS studies from 1994 to 2025. T stands for type and indicates whether it is Journal or Conference. J stands for Journal and C for Conference. The studies are sorted from oldest to newest. The Win Rate and Success Rate are expressed in percentage.}\label{tab:Sample FS studies}
\begin{tabular}
{|p{0.005\textwidth}|p{0.022\textwidth}|p{0.3\textwidth}|p{0.15\textwidth}|p{0.03\textwidth}|p{0.03\textwidth}|p{0.03\textwidth}|p{0.03\textwidth}|p{0.03\textwidth}|} 
\toprule
\textbf{T}&\textbf{Year}&\textbf{Title}&\textbf{Venue}&\textbf{\# Data sets}&\textbf{\# Base-lines}&\textbf{\# New Methods}&\textbf{Win Rate}&\textbf{Suc-cess Rate}\\\midrule
J &1994 &Using Mutual Information for Selecting Features in Supervised Neural Net Learning\cite{battiti_using_1994}&IEEE Transactions on Neural Networks [and Learning Systems] &2 &2 &1 &100 &- \\
J &2002 &Input feature selection for classification problems\cite{kwak_input_2002}&IEEE Transactions on Neural Networks [and Learning Systems]&2 &4 &2 &100 &50 \\
J &2005 &Feature selection based on mutual information criteria of max-dependency, max-relevance, and min-redundancy\cite{peng_feature_2005}&IEEE Transactions on Pattern Analysis and Machine Intelligence &4 &1 &1 &100 &- \\
J &2005 &MINIMUM REDUNDANCY FEATURE SELECTION FROM MICROARRAY GENE EXPRESSION DATA\cite{ding_minimum_2005}&Journal of Bioinformatics and Computational Biology &6 &2 &4 &100 &- \\
C &2006 &Conditional Infomax Learning: An Integrated Framework for Feature Extraction and Fusion\cite{lin_conditional_2006}&9th European Conference on Computer Vision &3 &4 &2 &100 &- \\
\hline
J &2013 &An Improved Minimum Redundancy Maximum Relevance Approach for Feature Selection in Gene Expression Data\cite{mandal_improved_2013}&Procedia Technology &4 &2 &1 &100 &- \\
C &2013 &Minimal-redundancy-maximal-relevance feature selection using different relevance measures for omics data classification\cite{yang_minimal-redundancy-maximal-relevance_2013}&IEEE Symposium on Computational Intelligence in Bioinformatics and Computational Biology (CIBCB) &5 &1 &2 &80 &100 \\
C &2014 &Compact Representation for Image Classification: To Choose or to Compress?\cite{zhang_compact_2014}&IEEE Conference on Computer Vision and Pattern Recognition &4 &2 &1 &50 &- \\
J &2015 &Feature selection using Joint Mutual Information Maximisation\cite{bennasar_feature_2015}&Expert Systems with Applications &11 &5 &2 &63.64 &- \\
J &2016 &The mRMR variable selection method: a comparative study for functional data\cite{berrendero_mrmr_2016}&Journal of Statistical Computation and Simulation &- &3 &2 &100 &100 \\
J &2017 &A Cosine-Similarity Mutual-Information Approach for Feature Selection on High Dimensional Datasets\cite{dubey_cosine-similarity_2017}&Journal of Information Technology Research &18 &1 &1 &72.22 &- \\
J &2017 &A new maximum relevance-minimum multicollinearity (MRmMC) method for feature selection and ranking\cite{senawi_new_2017}&Pattern Recognition &8 &2 &1 &50 &- \\
J &2017 &Fast-mRMR: Fast Minimum Redundancy Maximum Relevance Algorithm for High-Dimensional Big Data\cite{ramirez-gallego_fast-mrmr_2017}&International Journal of Intelligent Systems &5 &- &- &- &- \\
J &2017 &Maximum relevance minimum common redundancy feature selection for nonlinear data\cite{che_maximum_2017}&Information Sciences &4 &3 &1 &50 &- \\
J &2017 &Minimum redundancy maximum relevance feature selection approach for temporal gene expression data\cite{radovic_minimum_2017}&BMC Bioinformatics &3 &4 &2 &100 &- \\
J &2019 &Improved Measures of Redundancy and Relevance for mRMR Feature Selection\cite{jo_improved_2019}&Computers &10 &4 &1 &100 &- \\
C &2019 &Maximum Relevance and Minimum Redundancy Feature Selection Methods for a Marketing Machine Learning Platform\cite{zhao_maximum_2019}&IEEE International Conference on Data Science and Advanced Analytics (DSAA) &3 &5 &3 &66.67 &66.67 \\
J &2019 &On some aspects of minimum redundancy maximum relevance feature selection\cite{bugata_aspects_2019}&Science China Information Sciences &8 &- &- &- &- \\
\hline
J &2020 &A new feature selection algorithm based on relevance, redundancy and complementarity\cite{li_new_2020}&Computers in Biology and Medicine &15 &11 &1 &33.33 &- \\
J &2020 &Feature selection considering Uncertainty Change Ratio of the class label\cite{zhang_feature_2020}&Applied Soft Computing &14 &7 &1 &71.43 &- \\
J &2020 &Multilabel Feature Selection Using Relief and Minimum Redundancy Maximum Relevance Based on Neighborhood Rough Sets\cite{huang_multilabel_2020}&IEEE Access &9 &7 &1 &66.67 &- \\
J &2021 &A conditional-weight joint relevance metric for feature relevancy term\cite{zhang_conditional-weight_2021}&Engineering Applications of Artificial Intelligence &19 &7 &1 &63.16 &57.89 \\
J &2022 &Benchmark of filter methods for feature selection in high-dimensional gene expression survival data\cite{bommert_benchmark_2022}&Briefings in Bioinformatics &11 &- &- &- &- \\
J &2023 &CSCIM\_FS: Cosine similarity coefficient and information measurement criterion-based feature selection method for high-dimensional data\cite{yuan_cscim_fs_2023}&Neurocomputing &10 &9 &1 &70 &- \\
J &2023 &Feature Selection With Maximal Relevance and Minimal Supervised Redundancy\cite{wang_feature_2023}&IEEE Transactions on Cybernetics &12 &7 &1 &83.33 &- \\
J &2024 &Feature Selection Based on Intrusive Outliers Rather Than All Instances\cite{yuan_feature_2024}&IEEE Transactions on Image Processing &8 &10 &2 &62.25 &- \\
J &2024 &Minimising redundancy, maximising relevance: HRV feature selection for stress classification\cite{ihianle_minimising_2024}&Expert Systems with Applications &3 &4 &4 &100 &66.67 \\
J &2025 &Feature selection method based on wavelet similarity combined with maximum information coefficient\cite{yuan_feature_2025}&Information Sciences &9 &7 &1 &66.67 &- \\
\midrule\midrule
\multicolumn{4}{|l|}{Average (overall)} &7.8 &4.6 &1.6 &78.0 &73.6 \\
\midrule
\multicolumn{4}{|l|}{Average (1990s and 2000s)} &3.4 &2.6 &2.0 &100.0 &50.0 \\
\multicolumn{4}{|l|}{Average (2010s)}  &6.9 &2.9 &1.5 &75.7 &88.9 \\
\multicolumn{4}{|l|}{Average (2020s)}  &11.0 &7.7 &1.4 &68.5 &62.3 \\
\bottomrule
\end{tabular}
\end{table}

\section{Analysis of the Current Feature Selection Evaluation Landscape}\label{chap6:sec:Analysis of the Current Feature Selection Evaluation Landscape}

This section provides a critique of the current landscape of feature selection as found in the literature. The critique will focus on three main aspects: the choice of datasets, the choice of baselines, and the experimental setup.

\subsection{The Choice of Datasets} \label{chap6:subsec:The Choice of the Datasets}

The choice of datasets included in FS studies can introduce selection bias. Given the multitude of benchmark datasets used in FS studies, with sufficient computing resources, one can run experiments on many of them but include only those in which the proposed method outperforms the baselines. By analysing the sample studies, we observed that the number of datasets included in FS studies ranges from 2 \cite{battiti_using_1994,ihianle_minimising_2024} to 19 \cite{zhang_conditional-weight_2021} datasets.

Among the 28 sample FS studies analysed, only four included the nci9 dataset in their experiments \cite{ding_minimum_2005, peng_feature_2005, ramirez-gallego_fast-mrmr_2017, yuan_feature_2024}. nci9 is a complex multiclass (with 9 distinct classes) feature selection dataset from the National Cancer Institute (NCI) of the United States of America. 

The first two evaluations, including the nci9 dataset \cite{ding_minimum_2005, peng_feature_2005}, were reported over 20 years ago and reported error percentages. \citet{ramirez-gallego_fast-mrmr_2017} reported only the execution time, in seconds, for the sequential, parallel (for GPUs), and distributed (for Apache Spark) implementations of the mRMR algorithm. Only \citet{yuan_feature_2024}, published in 2024, reported the performance in accuracy of the nci9 dataset using LinearSVC. The highest reported accuracy score is 60\%, achieved by ReliefF \cite{yuan_feature_2024} for the number of features $k=300$. The second-highest accuracy is 58.33\%, achieved by their proposed method, Intrusive Outliers-based Feature Selection (IOFS), and its extreme variant (E-IOFS), for $k\in\{200, 250, 300\}$. 

Additionally, the average accuracy of the IOFS/E-IOFS method across all 8 datasets appearing in Table 7 in \citet{yuan_feature_2024} improves by up to 3.66 and 8.66 percentage points when nci9 and the three datasets with the lowest accuracy (``$^{-}$" marked) are excluded, respectively, from the calculation (see rows $\Delta$Avg (excluding nci9) and $\Delta$Avg (excluding ``$^{-}$" marked) in Table \ref{tab:aggregated accuracy from yuan_feature_2024 study} below).  

\underline{\textbf{Recommendations}}: Recommending which datasets should be included in feature selection (FS) studies is challenging, given the hundreds of benchmark datasets available in the literature and public repositories. Each dataset has its own characteristics and may be well-suited to certain use cases while being less relevant for others. An ideal scenario would be the availability of a public leaderboard that allows new methods evaluated on a given dataset to be compared with existing approaches without the need to retrain previously proposed methods.

\begin{table}[!htp]
\centering\small
\caption{Adaptation and extension of Table 7 from \citet{yuan_feature_2024}. Column ``Accuracy" represents the maximum accuracy by IOFS and E-IOFS. Column ``\#SelectedFeatures" represents the minimum number of selected features which attain the maximum reported accuracy. The three datasets with the lowest accuracy have been ``$^{-}$" marked.}
\label{tab:aggregated accuracy from yuan_feature_2024 study}
\begin{tabular}{l|r|r}\toprule
\textbf{Dataset} &\textbf{Accuracy} &\textbf{\#SelectedFeatures} \\\midrule
CLL\_SUB\_111$^{-}$ &73.87 &200 \\
Carcinom &97.13 &300 \\
ORL &96.25 &250 \\
Yale$^{-}$ &76.36 &150 \\
gisette &80.39 &300 \\
lung &96.55 &250 \\
lymphoma &92.71 &150 \\
nci9$^{-}$ &58.33 &200 \\
\midrule\midrule
Avg (across all datasets) &83.95 &225 \\
Avg (excluding nci9) &87.61 &229 \\
Avg (excluding ``$^{-}$" marked) &92.61 &250 \\
\midrule\midrule
$\Delta$Avg (excluding nci9) &3.66 &- \\
$\Delta$Avg (excluding ``$^{-}$" marked)  &8.66 &- \\
\bottomrule
\end{tabular}
\end{table}

\subsection{The Choice of Baselines} \label{subsec:The Choice of the Baselines}

To demonstrate the effectiveness of a new method, algorithm, or model, researchers typically need to choose at least one baseline to compare against. This is a crucial step because choosing the wrong baseline (by 'wrong', we mean a method that weakens the contribution of the new method) can undermine the contribution of the new method. Therefore, for a domain like feature selection with no standard baselines or a leaderboard, the choice of baselines can substantially affect the apparent performance advantage of a proposed method, especially when targeting top-tier venues similar to what is happening in other domains (e.g. tabular deep learning and data valuation in ML) as demonstrated in section \ref{chap6:sec:Introduction} above. However, it cannot be clearly identified as unethical because there is always a rationale for discarding one baseline or including another. This may also reflect perceived expectations within the peer-review process, which often seem to consider that high contribution equals beating [almost] all baselines. These biases led to many publications in which the proposed method almost always outperforms all baselines \cite{peng_feature_2005, mandal_improved_2013, berrendero_mrmr_2016, jo_improved_2019, ihianle_minimising_2024}. However, when those methods are included in benchmark studies by independent authors, the results are more mixed. If this course is not reversed or addressed, we might see an increasing number of publications with serious baseline selection bias but strong rationales.

Looking at the table \ref{tab:Sample FS studies} (see column ``Win Rate"), we can observe that the win rate of a new method against the competing methods ranges from 33.3\% to 100\%. In fact, as the number of datasets or baselines increases, the win rate decreases (see section \ref{chap6:sec:Lessons Learned} for a more detailed analysis).
In general, the fewer the datasets and the competing methods, the higher the win rate of the new method.

\underline{\textbf{Recommendations}}: When developing a new feature selection method, we argue that the all-feature approach (i.e. using the complete original feature set) and KBest should be considered essential baselines. The all-feature performance serves as a key reference point that any feature selection (FS) method aims to outperform; it is fundamental for assessing whether feature selection has been successful. KBest, on the other hand, is one of the simplest and most computationally efficient feature selection methods, making it an important benchmark against which new FS approaches should be evaluated. More broadly, in machine learning, it is widely recommended to begin with simple methods before progressing to more complex ones \cite{guyon_introduction_2003}, reinforcing the need to compare new FS techniques against KBest.

\subsection{The Experimental Setup} \label{subsec:The Experimental Setup}

The experimental setup is another important step that can both positively and negatively impact the effectiveness of a given FS method. For instance, if the number of features $k$ is a hyperparameter, setting an appropriate value of $k$ will directly determine how well the method performs on the datasets used in the experiments. Another important decision is the model to include in the experiment, as different models or model families learn differently and yield different outcomes on a given task. To ensure the method is not biased or overly dependent on a particular model, one should include a variety of models with diverse learning principles. If a new method is model-dependent, the authors should clearly state this in their study.

Some studies manually set $k$ to a single value without providing a rationale for that choice. Others give a set of values of $k$ (e.g. $k\in\{50, 100, 150, 200, 250, 300\}$) with a number $n_k$ ($n_k>1$) of features added at a time. These two approaches are not optimal because we know that even the slightest update of the set of selected feature $\mathcal{S}$ (e.g. one added to or removed from $\mathcal{S}$) can highly impact the performance of the selected features. 
This is because the predictive performance of $\mathcal{S}$ is determined by the individual or non-contextual relevance of each feature as well as the interactive or contextual relevance (i.e. redundancy and complementarity) \cite{guyon_introduction_2003}. While the non-contextual relevance is constant, the contextual relevance changes every time $\mathcal{S}$ is modified. 

\underline{\textbf{Recommendations}}: Most feature selection methods take in $k$ as a hyperparameter. For those, one should systematically select the best value of $k$ for $k\in[2;100]$ (for example) and $n_k=1$. This guarantees that the feature selection task returns the minimum $k$ with the highest accuracy. This is not guaranteed if $n_k\geq2$ or if the search starts at a higher value (for instance, $k\in[10;100]$ or $k\in[50;80]$). Some methods have additional hyperparameters that need to be finetuned. For instance, IOFS and E-IOFS \cite{yuan_feature_2024} have a hyperparameter $\alpha\in [0;1]$. The authors have demonstrated in their study that for a given dataset, a specific value of $\alpha$ is needed to achieve the highest accuracy. Therefore, employing such a FS method without any hyperparameter finetuning can be misleading.

\section{Lessons Learned} \label{chap6:sec:Lessons Learned}

This section summarises the lessons learned from our study of the 28 filter feature selection studies.

\subsection{Extracted Data Statistics}

The overall win rate is 78\% in table \ref{tab:Sample FS studies} (see row ``Average (overall)"). The average win rate per decade is 100\%\footnote{We combined the 1990s and 2000s because there is only one study from the 1990s in the sample studies.},
75.7\%, and 68.5\%, respectively for the 1990s and 2000s, 2010s, and the 2020s. This means that the win rate of studies published in those different decades has been decreasing from one decade to the next.

However, this can be explained by the increasing number of datasets and baselines, and the decreasing number of new methods included in those studies (see section \ref{chap6:subsec:Univariate, Bivariate and Trivariate Correlations} for more details). The decreasing of the win rate from one decade to the next can also be explained by the fact that early in the field, it was easier to develop new methods that outperform the baselines. In a mature field, it is harder to develop new methods that outperform the baselines.


Regarding the average success rate, only 21.4\% (6 out of 28) of the examined studies reported all-feature results. The success rate first increases from 50\% in the 1990s and 2000s to 88.9\% in the 2010s, then decreases to 62.3\% in the 2020s. These statistics are less reliable because the results of all-feature is under-reported, as mentioned above.

Overall, the average number of datasets and baselines has increased over the decades, while the average number of new methods has decreased. The average win rate has also decreased over recent decades as the domain has matured and the average number of datasets and baselines has increased.

The only unexpected finding is that in the majority of scenarios (Average (overall), Average (1990s and 2000s), and Average (2020s)), the average win rate is greater than the average success rate; meaning that, on average, \textbf{1) most new filter feature selection methods outperform multiple baselines (1 to 11 of them) but are unable to outperform the original dataset (all-feature)}; this is the first lesson learned.

\subsection{Univariate, Bivariate and Trivariate Correlations}\label{chap6:subsec:Univariate, Bivariate and Trivariate Correlations}

This section discusses the resulting insights of the data analysis described in section \ref{chap6:subsec:Data Analysis} above.

\begin{figure}[!htp]
\centering
\includegraphics[width=0.99\textwidth]{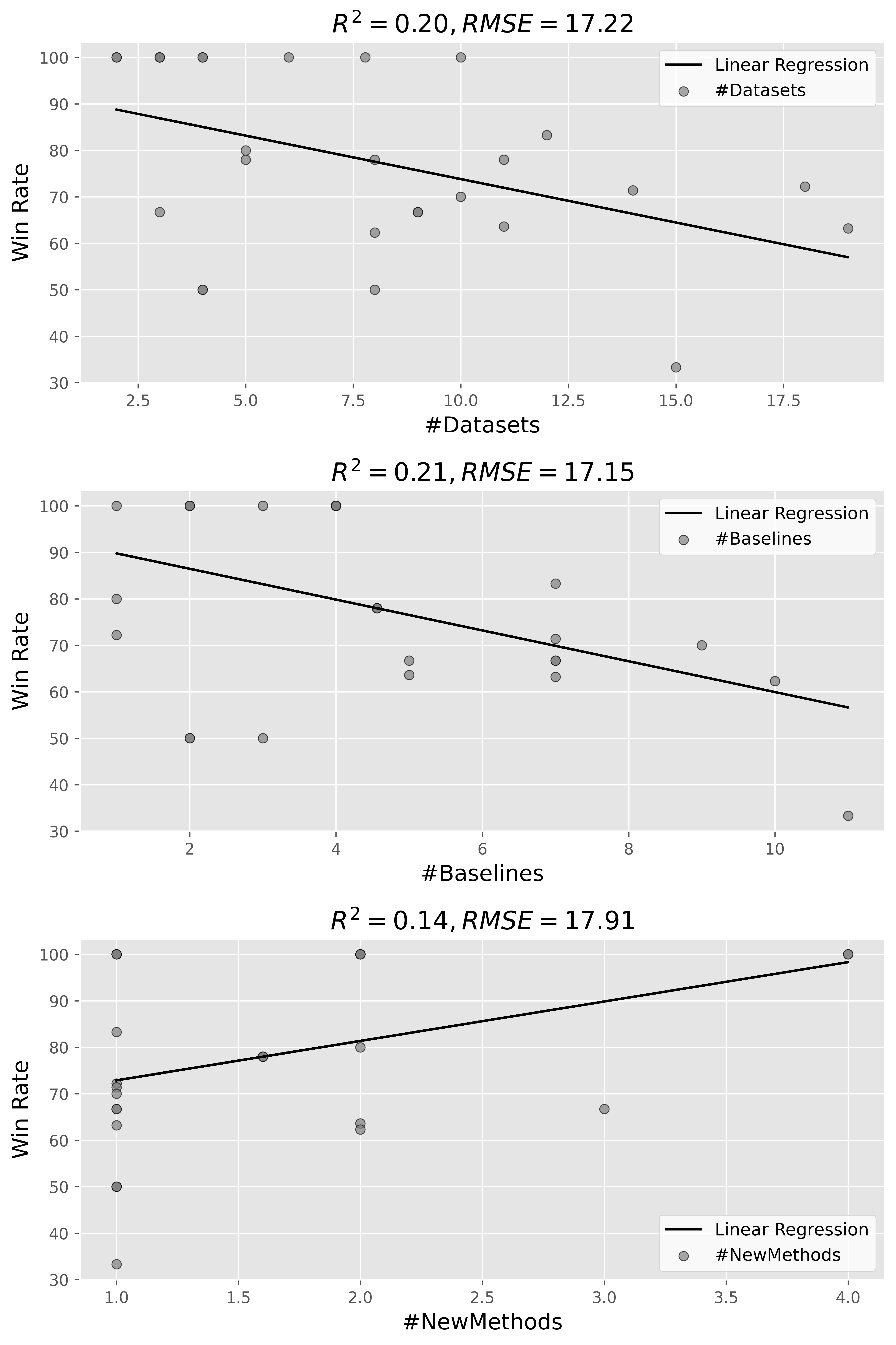}
\caption{Univariate correlation between Win Rate, on one side, and the number of datasets (\#Datasets), the number of baselines (\#Baselines), and the number of new methods (\#NewMethods), on the other side.}
\label{chap6:fig:univariate correlation between Win Rate and Sample data}
\end{figure}

\textbf{Univariate Correlations.} Figure \ref{chap6:fig:univariate correlation between Win Rate and Sample data} presents the correlation of win rate with each of the three extracted variables, i.e. number of datasets (\#Datasets), number of baselines (\#Baselines), and number of new methods (\#NewMethods). \textbf{Win rate is negatively related to the number of datasets and baselines, but positively related to the number of new methods.} In other words, win rate decreases as either the number of datasets or baselines increases. However, win rate increases as the number of new methods increases. Win rate is influenced in decreasing order by the number of datasets, baselines, and new methods, as expressed by $R^2$ and RMSE (see figure \ref{chap6:fig:univariate correlation between Win Rate and Sample data}).

\begin{figure}[!htp]
\centering
\includegraphics[width=0.99\textwidth]{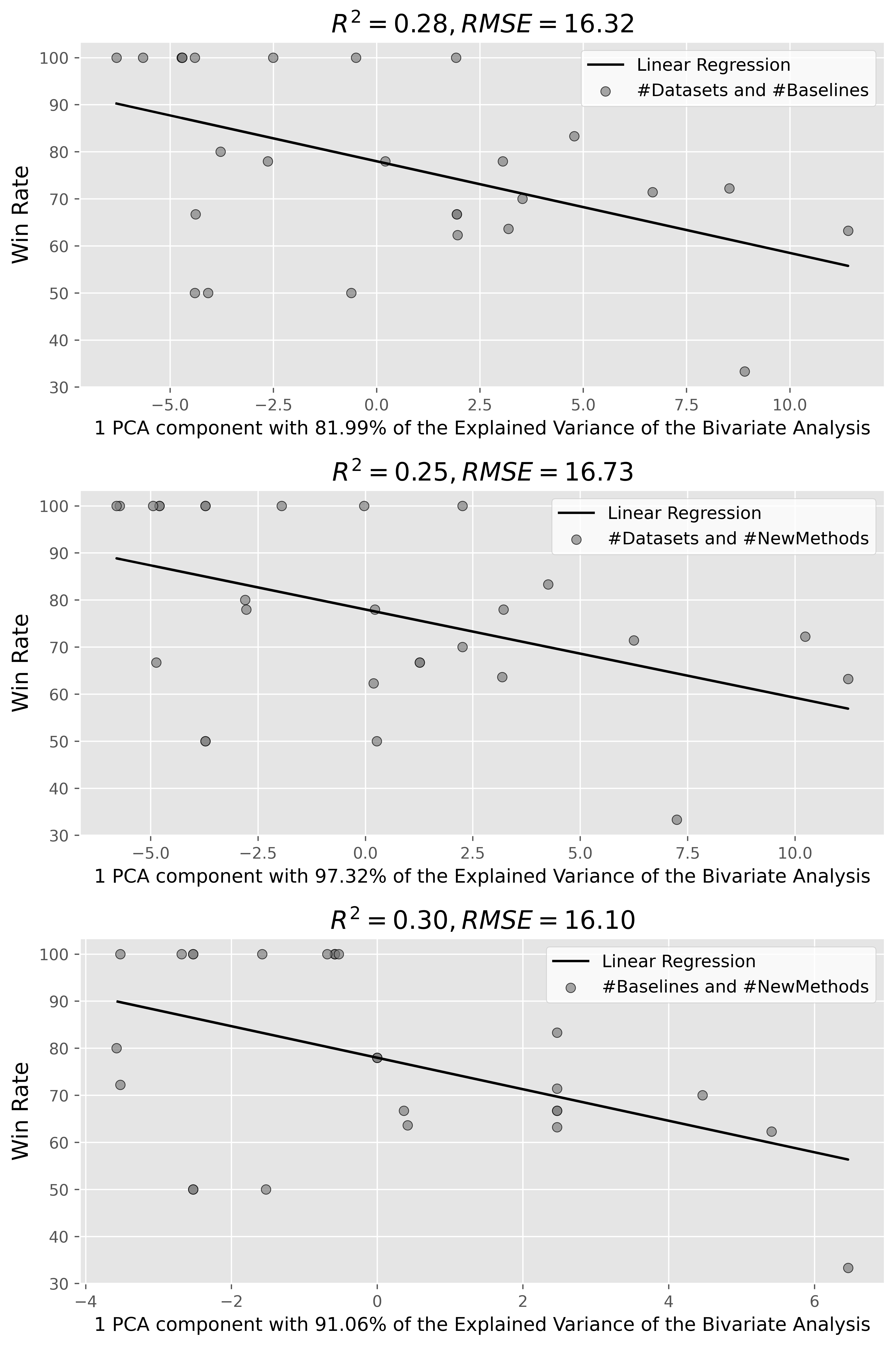}
\caption{Bivariate correlation between Win Rate, on one side, and the number of datasets (\#Datasets), the number of baselines (\#Baselines), and the number of new methods (\#NewMethods), on the other side.}
\label{chap6:fig:bivariate correlation between Win Rate and Sample data}
\end{figure}

\textbf{Bivariate Correlations.} Figure \ref{chap6:fig:bivariate correlation between Win Rate and Sample data} displays the correlation of win rate with two of the three extracted variables simultaneously. \textbf{Win rate is negatively related to any combination of two out of the three extracted variables.} This means that win rate decreases as any two of the three extracted variables simultaneously increase. Win rate is more dependent simultaneously on \#Baselines and \#NewMethods, \#Datasets and \#Baselines, and \#Datasets and \#NewMethods, respectively, as expressed by $R^2$ and RMSE (see figure \ref{chap6:fig:bivariate correlation between Win Rate and Sample data}). The bivariate analysis captures the data better than the univariate as it increases $R^2$ and decreases RMSE.

\begin{figure}[!htp]
\centering
\includegraphics[width=0.99\textwidth]{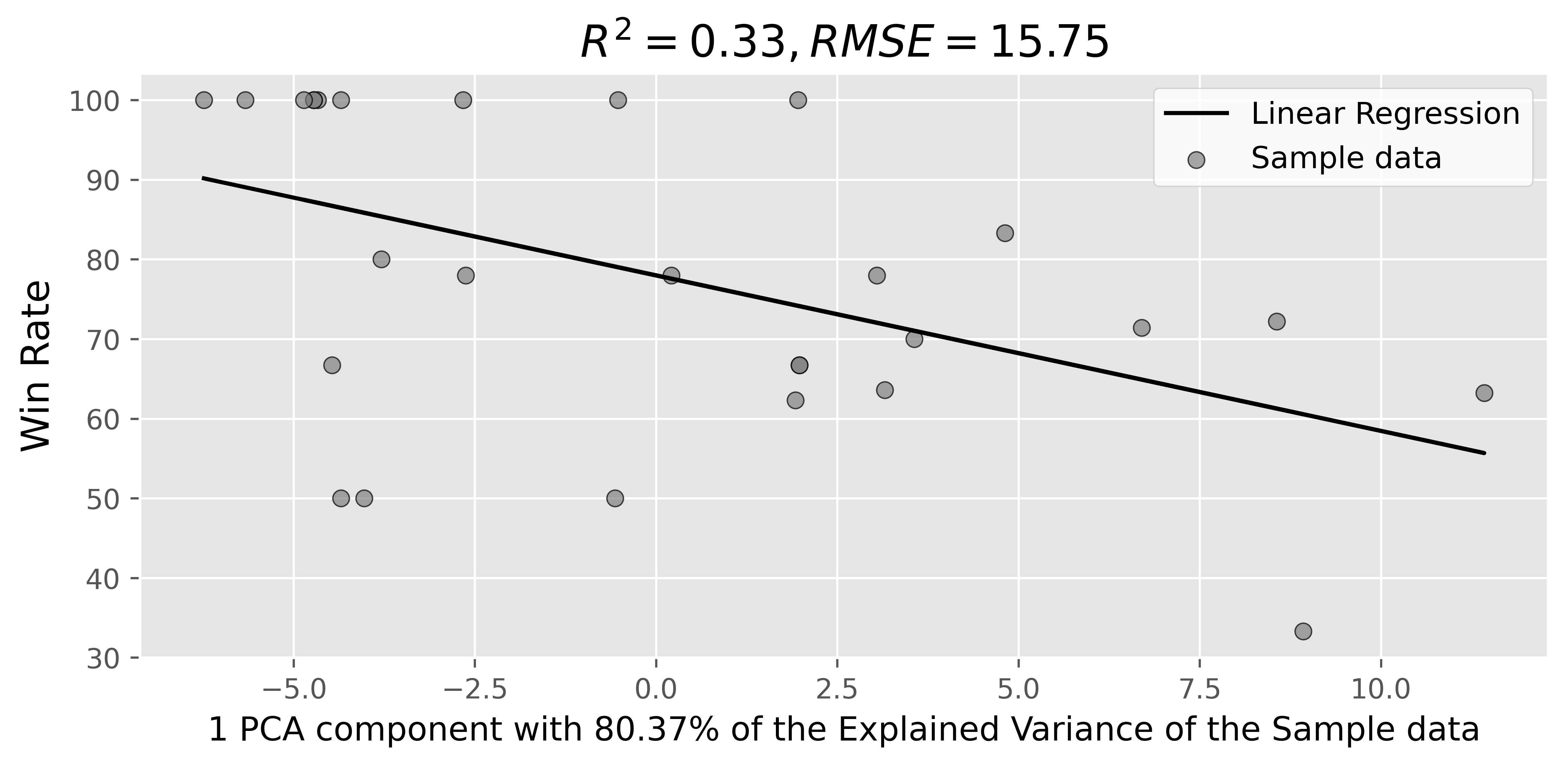}
\caption{Trivariate correlation between Win Rate, on one side, and the number of datasets (\#Datasets), the number of baselines (\#Baselines), and the number of new methods (\#NewMethods), on the other side.}
\label{chap6:fig:correlation between Win Rate and Sample data}
\end{figure}

\textbf{Trivariate Correlations.} Figure \ref{chap6:fig:correlation between Win Rate and Sample data} highlights the correlation of win rate with the three extracted variables simultaneously. \textbf{Win rate is negatively related to the three extracted variables, all taken together.} This means that win rate decreases as all three extracted variables simultaneously increase.
The trivariate analysis captures the data better than the bivariate, as it improves $R^2$ and RMSE.

\textbf{The lessons learned here are: 2) Win rate is negatively related to the number of datasets (\#Datasets) and the number of baselines (\#Baselines), but positively related to the number of new methods (\#NewMethods); 3) Win rate is negatively related to any combination of two out of the three extracted variables; and 4) Win rate is negatively related to the three extracted variables all taken together.} The relationship among the number of datasets (\#Datasets), the number of baselines (\#Baselines), and the number of new methods (\#NewMethods) is more complex than the analysis presented above. The relationship not only depends on the numbers, but also on the type of datasets (high-dimensional versus medium-to-low-dimensional, or biological data versus images or text data, discrete versus continuous data), the type of baselines (information theory-based versus heterogeneous baselines including embedded or wrappers). However, our analysis gives a broad idea of the win rate trend as the number of datasets (\#Datasets), the number of baselines (\#Baselines), and the number of new methods (\#NewMethods) increase.

\section{General Recommendations}\label{chap6:sec:General Recommendations}

This section provides two general recommendations. These are complementary to the specific ones given in sections \ref{chap6:subsec:The Choice of the Datasets}, \ref{subsec:The Choice of the Baselines}, and \ref{subsec:The Experimental Setup} above.

\subsection{Recommendation 1: Study of the Factors that Influence Feature Selection} \label{chap6:subsec:Recommendation 1: Study of the Factors that Influence Feature Selection}

Despite efforts over the years to develop new filter feature selection methods, there is a limited, up-to-date understanding of the factors that influence feature selection methods, especially the relevance and redundancy estimators used by those methods. For instance, \citet{strobl_bias_2007} showed that the RF model is biased towards features with high cardinality. This influences the feature relevance estimation using methods such as Gini importance or SHAP values. \citet{loecher_debiasing_2022} and \citet{baudeu_are_2023} showed that the bias towards features with high cardinality extends to other tree-based models as well. They argued that it is an inherent property of Decision Trees. However, there have been many attempts to address that bias \cite{loecher_debiasing_2022, baudeu_are_2023}. The discussion of those attempts and their effectiveness is beyond the scope of this study.

Cardinality also influences information theory-based relevance estimators, particularly entropy (directly) and mutual information (indirectly through entropy). However, this influence is heavily dependent on the probability distribution of the values. It is more noticeable when the values are [fairly] uniformly distributed (the notion of maximum entropy in information theory: it increases with the cardinality if and only if the distribution is uniform).

The work presented by \citet{ebiele_impact_2024} shows that the coefficient of variation (CV) also influences the relevance estimations of the features. \citet{ebiele_impact_2024} do not only focus on tree-based models (RF and XGB) like most studies in this research area, but also includes linear models (Logistic Regression), probabilistic models (Naive Bayes), and [simple] neural networks (MLP).

Cardinality and coefficient of variation are not the only factors that influence relevance estimation and feature selection. However, these factors are underexplored. For instance, what is the impact on relevance estimation and feature selection when cardinality and coefficient of variation are both high?

\subsection{Recommendation 2: Feature Selection is an Exploratory Task, so Explore}
There have been a considerable number of new feature selection methods or extensions of existing ones \cite{li_feature_2017, theng_feature_2024}. However, limited efforts have been made to better understand or use them. For instance, scholars and practitioners still do not know when to use the difference or the quotient formulation of mRMR; as the dataset, relevance estimator, and even the experimental setup vary, one formulation outperforms the other, and vice versa \cite{berrendero_mrmr_2016, ding_minimum_2005}. This means that, to be sure mRMR is not suitable for a given feature selection task, one needs to try both, potentially with different relevance and redundancy estimators. However, in the literature, very few studies have investigated both formulations when applying mRMR to feature selection. For example, only 7 among the 28 sample studies investigated both formulations of mRMR \cite{ding_minimum_2005, mandal_improved_2013, jo_improved_2019, zhao_maximum_2019, yuan_cscim_fs_2023, ihianle_minimising_2024, yuan_feature_2025}.

\emph{Recommendation 2.1:} FFS method evaluation studies should explore both the difference and the quotient formulation of mRMR for each dataset.

\emph{Recommendation 2.2:} FFS method evaluation studies should explore a range of different relevance and redundancy estimators for each dataset.

Another thing that can be observed is the limited finetuning of the hyperparameters of the feature selection methods. For instance, the number $k$ of features to be selected is a hyperparameter in the majority of cases, yet in most studies, $k$ is set to a single value. This is relevant because some methods are excellent at selecting fewer features, while others do better as the number $k$ increases. 
Some methods have additional hyperparameters that need to be finetuned as well. For instance, IOFS \cite{yuan_feature_2024} has a hyperparameter $\alpha\in [0;1]$. The authors have demonstrated that for a given dataset, a specific value is needed to achieve the highest possible accuracy.

\section{Conclusion}\label{chap6:sec:Summary and Conclusion}

This study analysed a sample of 28 studies from 1994 to 2025. The studies were critiqued with respect to three main aspects: the choice of the baselines, the choice of the datasets, and the experimental setup. These aspects are important because they define the context in which the significance of new FFS methods is examined. After each critique, specific recommendations for future evaluations have been provided. Four findings were made examining the data extracted from those studies: 1) Most new filter feature selection methods outperform multiple baselines (1 to 11 of them) but are unable to outperform the original dataset (all-feature); 2) Win rate is negatively related to the number of datasets and the number of baselines, but positively related to the number of new methods; 3) Win rate is negatively related to any combination of two out of the three extracted variables; and 4) Win rate is negatively related to the three extracted variables when they are all taken together. Finally, the study makes two general recommendations for further work on the feature selection task: Recommendation 1)-Researchers and practitioners should make more effort to study the factors (see section \ref{chap6:subsec:Recommendation 1: Study of the Factors that Influence Feature Selection}) that influence feature selection, and Recommendation 2)-feature selection is an exploratory task, so researchers and practitioners should explore different configurations of data and methods when performing a feature selection task.

The analysis of the extracted data suggests that win rate decreases as the number of datasets (\#Datasets), the number of baselines (\#Baselines), and the number of new methods (\#NewMethods) increase individually (except \#NewMethods), two at a time, or all three together. This means that if a study reports a large number of datasets, baselines, and new methods while presenting their new methods as clearly superior, it should raise a question of selection bias in the choice of the datasets or baselines, or in the experimental setup.

For future work, the study selection could be systematised through keyword search in top databases (e.g., IEEE, ACM, Web of Science) and extended to wrapper and embedded methods as well as application studies which do not introduce any new method. This will considerably increase the sample size. The data extraction and analysis procedures presented in this study will be repeated. Finally, Large language models (LLMs) will also be explored as tools to assist with data extraction. 


\backmatter

\bmhead{Supplementary information} The data extracted from the 28 sample studies and the data analysis code files are available at \url{https://github.com/malick-jaures/FFS_Meta_Analysis.git}.

\bmhead{Acknowledgements} This research was conducted with the financial support of Taighde E\'ireann – Research Ireland under Grant Agreement No. 13/RC/2106\_P2 at the ADAPT Centre at University College Dublin.

\section*{Declarations}
\textbf{Declaration on Generative AI.} During the preparation of this work, Grammarly Premium was used for grammar and spelling checks. No texts have been generated using large language models (LLMs). No prompt engineering was performed to improve the tone.

\bibliography{sn-bibliography}

\end{document}